\documentclass[11pt]{article}

\usepackage[preprint]{acl}

\usepackage{times}
\usepackage{latexsym}
\usepackage[T1]{fontenc}
\usepackage[utf8]{inputenc}
\usepackage{microtype}
\usepackage{inconsolata}
\usepackage{graphicx}
\usepackage{booktabs}
\usepackage{colortbl}
\usepackage{amsmath,amssymb}

\usepackage[accsupp]{axessibility}  

\definecolor{oursblue}{RGB}{232,244,255}

\begin{document}
\title{An Efficient Streaming Video Understanding Framework with Agentic Control}

\author{
\textbf{Jinming Liu}$^{1,2,*}$ \quad
\textbf{Jianguo Huang}$^{1,2,*}$ \quad
\textbf{Zhaoyang Jia}$^{3}$ \quad
\textbf{Jiahao Li}$^{3}$ \\
\textbf{Xiaoyi Zhang}$^{3}$ \quad
\textbf{Zongyu Guo}$^{3}$ \quad
\textbf{Bin Li}$^{3}$ \quad
\textbf{Wenjun Zeng}$^{2}$ \quad
\textbf{Yan Lu}$^{3}$ \quad
\textbf{Xin Jin}$^{2}$ \\[0.6em]
$^1$Shanghai Jiao Tong University \\
$^2$Eastern Institute of Technology, Ningbo, China \quad
$^3$Microsoft Research Asia
}

\maketitle
\renewcommand\thefootnote{\fnsymbol{footnote}}
\footnotetext[1]{Equal contribution.}

\begin{abstract}
Streaming video requires handling dynamic information density under strict latency budgets. Yet, existing methods typically employ static strategies, like fixed memory compression or single-model reliance, forcing a trade-off: fast models fail on complex queries, while `always-on' heavy models violate real-time constraints and overcomplicate some simple queries.
Rather than fixing these decisions upfront, we propose R3-Streaming (Remember, Respond, Reason), which formulates streaming video understanding as a cascaded control problem: for each query, the system compresses memory, judges response readiness, and routes computation sequentially, so that each downstream decision builds on progressively refined information states. To optimize this pipeline, we introduce an age-aware forgetting policy for memory compression, as aggressively compressing historical frames can yield substantial performance gains. For compute routing, we propose TB-GRPO, a target-balanced reinforcement learning (RL) objective that routes hard queries to a stronger model while preventing mode collapse.
Extensive evaluations demonstrate that R3-Streaming achieves state-of-the-art results among streaming MLLMs, reaching 57.92 on OVO-Bench and 76.36 on StreamingBench, all while reducing visual token usage by 95\%–96\%.
\end{abstract}

\section{Introduction}
Recent video-language models perform well on offline video understanding, including long-context tasks such as captioning, summarization, and question answering \cite{mvbench2024,mlvu2024,longvideobench2024,longvu2024}. However, most of these systems process full clips and are hard to deploy in real-time streams, where frames arrive continuously and each decision must satisfy strict latency constraints.

Existing streaming methods usually optimize one part of the pipeline at a time, such as response triggering or token/frame compression \cite{videollmonline2024,timechatonline2025,streamingassistant2025,proactivevideoqa2025,streamingvlm2025}. While these methods improve efficiency, they often hit a performance bottleneck where answer quality degrades sharply under strict latency constraints. This sub-optimal trade-off arises because memory retention, response timing, and reasoning depth are not dynamically coordinated, and therefore fail to adapt to the varying informational density of real-time streams.

Therefore, we reframe streaming video understanding not as a passive, feed-forward task, but as an agentic control problem. A streaming agent can autonomously manage its own state and dynamically decide its compute path through three sequentially coupled decisions per turn: (i) Remember, how much historical evidence should remain in memory; (ii) Respond, whether the current context supports a reliable answer; and (iii) Reason, whether the query should be solved by an advanced slow model.

\begin{figure*}[t]
  \centering
  \includegraphics[width=\textwidth]{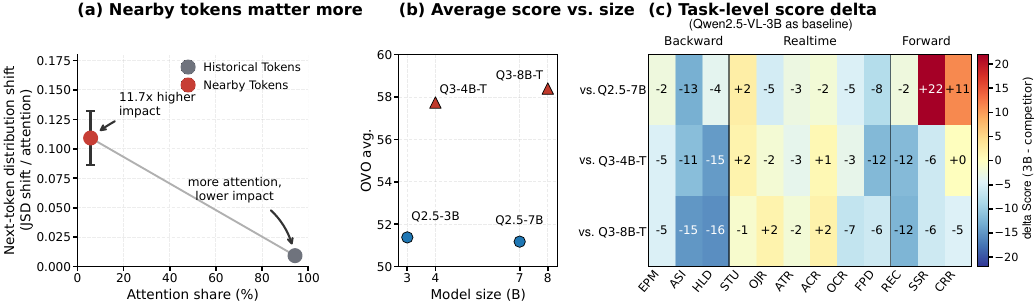}
  \vspace{-6mm}
  \caption{Empirical motivations for R3-Streaming. (a) Historical tokens receive most of the visual attention, yet deleting them has only a minor impact on the output token distribution. In contrast, removing nearby tokens induces much larger shifts in the next-token distribution. (b) Large models are not always better, where Q2.5/Q3 = Qwen2.5/3-VL, T = Thinking. (c) Per-task performance differences on OVO-Bench between Qwen2.5-VL-3B and larger models. Positive values indicate tasks where the 3B model performs better.}
  \vspace{-2mm}
  \label{fig:teaser_intro}
\end{figure*}

This formulation is motivated by two empirical findings (Sec.~\ref{sec:prelim_findings}), as illustrated in Fig.~\ref{fig:teaser_intro}. 
First, in Fig.~\ref{fig:teaser_intro}a, we attempt to analyze the impact of removing historical tokens or nearby tokens on the final output distribution (measured by Jensen--Shannon divergence, JSD~\cite{lin1991divergence,kim-etal-2020-interpretation}), and find that historical tokens receive most of the visual attention, yet removing them has far less impact on the output token distribution than removing nearby tokens. This suggests that historical tokens contain substantial redundancy, which can induce misleading attention allocation away from more informative nearby tokens. Fig.~\ref{fig:finding1_combined} further illustrates this point — we find that aggressively compressing historical tokens can actually improve benchmark performance.
Second, model-scale gains are non-monotonic: small models may outperform large thinking models on specific streaming tasks (Fig.~\ref{fig:teaser_intro}b,c), so always invoking heavy "thinking" inference is not optimal.

Based on these findings and the formulation, we propose \textbf{R3-Streaming} (\textbf{R}emember, \textbf{R}espond, \textbf{R}eason), a three-decision policy system with: (i) Active Forgetting, a training-free age-aware memory compressor that reduces stale context while preserving recent frames at high fidelity; (ii) Proactive Response, a readiness head that emits \texttt{<Routine>} and defers answering when evidence is insufficient; and (iii) Adaptive Thinking, a routing policy where the fast model either answers directly or emits \texttt{<Escalate>} to invoke a heavier reasoning model for compute routing. The resulting framework is summarized in Fig.~\ref{fig:r3_framework}. Among the three decisions, learning the binary routing policy in Adaptive Thinking poses the greatest optimization challenge: standard reinforcement learning (GRPO~\cite{deepseekmath2024}) and AutoThink methods~\cite{learnwhenthink2025} often struggle with stability, frequently leading to uncontrollable or excessively high escalation rates. The resulting policy may still invoke the heavy reasoning path at a frequency that exceeds the available compute budget. We address this with TB-GRPO, a target-balanced RL objective that stabilizes routing and keeps escalation rates within deployable compute budgets. Our contributions are:
\begin{enumerate}
  \item We establish two findings for streaming VLMs: important signal is strongly recent-focused, while historical tokens are often redundant and can lead to misleading attention; model-scale gains are non-monotonic across streaming tasks, motivating selective reasoning instead of always-on thinking.
  \item We introduce R3-Streaming, a cascaded control framework whose core technical contributions are an age-aware forgetting policy for memory compression and TB-GRPO, a target-balanced RL objective that stabilizes compute routing within deployable escalation budgets.
  \item We validate the framework on streaming benchmarks, reaching 57.92 on OVO-Bench and 76.36 on StreamingBench with 95\%--96\% visual-token reduction. On StreamingBench, our adaptive routing outperforms direct slow-only inference, achieving a better efficiency--accuracy trade-off.
\end{enumerate}

\section{Related Works}
\label{sec:related_work}
\paragraph{\textbf{Online and long-video VLMs.}}
Streaming VLMs process continuously arriving frames with state updates, token pruning, cache maintenance, or decision-reaction modules \cite{videollmonline2024,timechatonline2025,livevlm2025,dispider2025,streamingvlm2025}. In parallel, long-video VLMs and benchmarks study temporal reasoning and memory/compression over extended clips \cite{mvbench2024,mlvu2024,longvideobench2024,hiermem2024,longvu2024}. Recent online benchmarks such as StreamingBench, OVBench, and OVO-Bench further expose timestamped understanding and reasoning challenges \cite{streamingbench2024,ovbench2025,ovobench2025}. However, most prior systems optimize memory retention, response timing, or answer quality separately. R3 instead treats streaming understanding as a joint control problem over what to remember, when to respond, and when to invoke stronger reasoning.

\paragraph{\textbf{Adaptive reasoning and routing.}}
Adaptive decision methods study when models should deliberate versus answer directly. AutoThink uses multi-stage RL to learn when additional thinking is useful \cite{learnwhenthink2025}, and AdaptThink trains dynamic switching policies between thinking and no-thinking modes \cite{adaptthink2025}. These methods are not designed around strict streaming budgets, where excessive slow-model calls directly harm deployability. Our TB-GRPO routing objective adds target-band control over escalation frequency, preserving answer quality while keeping slow-path usage predictable.

\section{Preliminary Findings}
\label{sec:prelim_findings}
Before introducing module details, we summarize two empirical findings that motivate our design. 

\begin{figure}[t]
  \centering
  \includegraphics[width=\linewidth]{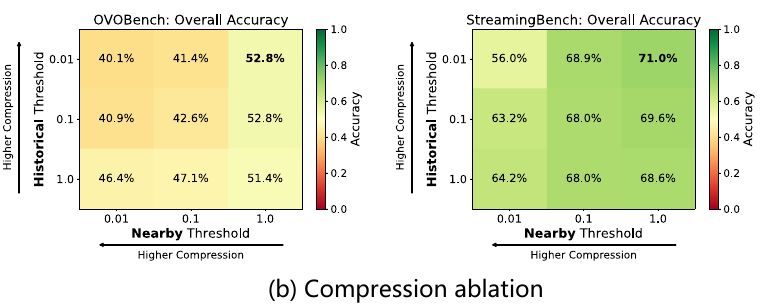}
  \vspace{-5mm}
  \caption{Compression threshold ablations on OVO-Bench and StreamingBench. The results show that preserving nearby context while compressing history gives the best performance. Refer to Appendix~\ref{sec:supp_backbone_grids} for results across additional models and benchmarks.}
  \vspace{-2mm}
  \label{fig:finding1_combined}
\end{figure}

\paragraph{\textbf{Finding 1: Important signal is strongly concentrated in recent context, while historical tokens often contain harmful redundancy and can induce misleading attention allocation.}}
Fig.~\ref{fig:teaser_intro}a and Fig.~\ref{fig:finding1_combined} summarize the evidence for Finding 1. The deletion analysis on OVO-Bench in Fig.~\ref{fig:teaser_intro}a shows that Nearby Tokens cause much larger next-token distribution shifts than Historical Tokens, even though Historical Tokens receive most of the attention mass. This indicates that important signal is concentrated in nearby context, whereas historical memory may contribute noisy signals.

\begin{figure}[t!]
  \centering
  \includegraphics[width=\linewidth]{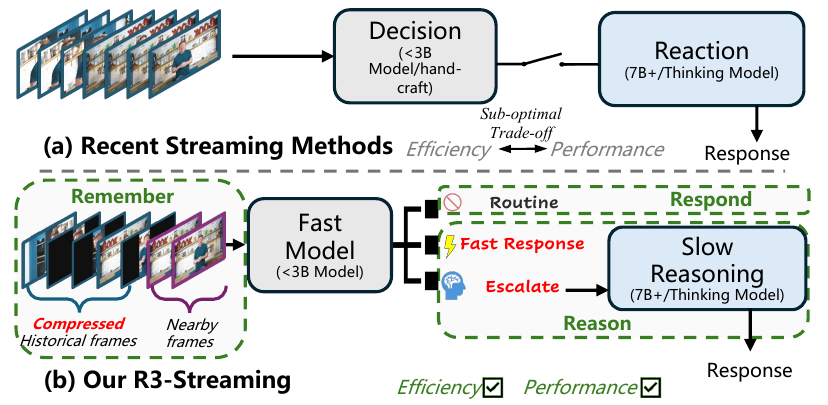}
  \vspace{-5mm}
  \caption{Framework comparison. (a) Recent decision-reaction streaming methods rely on a single reaction model, leading to a sub-optimal efficiency-performance trade-off. (b) R3-Streaming decomposes each streaming step into three cascaded decisions—memory compression (Remember), readiness estimation (Respond), and compute routing (Reason).}
  \label{fig:r3_framework}
\end{figure}

As a downstream benchmark check in Fig.~\ref{fig:finding1_combined}, we observe a clear asymmetry: preserving \textbf{Nearby} context is critical, while \textbf{Historical} memory can be compressed more aggressively. On StreamingBench, the best setting (Nearby$=1.0$, Historical$=0.01$) reaches 71.0, which is +2.4 over no compression. In contrast, once nearby context is heavily compressed, performance drops much. This indicates that compressing historical tokens while retaining nearby tokens can mitigate the misleading attention and improve model performance.

\paragraph{\textbf{Finding 2: Model-scale gains are non-monotonic across streaming tasks.}}
As shown in Fig.~\ref{fig:teaser_intro}b, performance on streaming tasks does not scale monotonically with parameter size. Small fast models can match or outperform larger models on specific task groups. Notably, the Qwen2.5-VL-3B model exceeds the 7B model, while the Qwen3-VL-4B-Thinking model is comparable to Qwen3-VL-8B-Thinking. This counter-intuitive trend is further elucidated by the task-level deltas in the same panel. The 3B instruct model demonstrates superior performance over larger, thinking-enabled models specifically within the Realtime and Forward task groups. Such strong task heterogeneity indicates that 'always-on' heavy inference is not only compute-intensive but can also be sub-optimal for specific streaming perception tasks. 

\paragraph{\textbf{Implication for R3 design.}}
These findings map directly to the R3 design. \textbf{Finding 1} motivates \textbf{Remember}: Active Forgetting keeps recent context at high fidelity and compresses old memory aggressively. \textbf{Finding 2} motivates \textbf{Reason}: Adaptive Thinking routes only hard queries to slow/thinking inference instead of using an always-on large model. \textbf{Respond} is inspired by recent streaming decision-reaction frameworks \cite{dispider2025,streamagent2025}, and is implemented here as a lightweight readiness head on the fast model to decide whether to answer now or defer. 

\section{Method}
\label{sec:method}

\subsection{Overview and Problem Formulation}
At each streaming step $t$, the system observes stream history $x_{1:t}$, receives query $q_t$, maintains a memory state $M_t$, and faces a chain of increasingly refined decisions (Fig.~\ref{fig:r3_framework}). First, the arriving frame must be assimilated into a compact memory state $M_t$ without drowning subsequent reasoning in stale context (\textbf{Remember}). Given the updated memory, the system then evaluates whether the accumulated evidence already grounds a reliable answer or whether deferral is more prudent (\textbf{Respond}). Only when readiness is confirmed does the system commit compute---choosing between a lightweight direct path and a heavier reasoning path based on estimated query difficulty (\textbf{Reason}).

Because each downstream decision consumes the output of the preceding one, errors compound: a noisy memory misleads readiness estimation, which in turn triggers unnecessary escalation. This cascaded dependency motivates a coordinated system design rather than treating memory, readiness, and routing as unrelated decisions.

Formally, we model decision-making with the action space $A = \{\texttt{<Answer>},\ \texttt{<Escalate>},\ \texttt{<Routine>}\}$, where \texttt{<Answer>} returns grounded response $y_t$, \texttt{<Escalate>} delegates the query to a slow/thinking model, and \texttt{<Routine>} defers output when current evidence is insufficient. The objective is to maximize answer quality while minimizing latency and unnecessary slow-model calls under streaming causality. The following subsections detail each decision stage in the order it executes.


\subsection{Remember (Active Forgetting)}
Streaming history grows linearly, yet the signal that actually matters for the next answer concentrates overwhelmingly in recent frames (Sec.~\ref{sec:prelim_findings}). This asymmetry makes a uniform compression schedule wasteful: it either discards recent context or retains irrelevant history that introduces attentional noise. We therefore partition the streaming history $x_{1:t}$ into two segments based on a temporal window $W$:
\begin{equation}
\begin{aligned}
M_t ={}& \text{Compress}(x_{t-W+1:t}, \tau_{near})\\
&{}\cup \text{Compress}(x_{1:t-W}, \tau_{hist}),
\end{aligned}
\end{equation}
where $\tau_{near}$ and $\tau_{hist}$ denote the compression thresholds for nearby and historical zones, respectively. 

Guided by the observation that nearby context is critical for accuracy while stale history introduces noise and misleading attention (Sec.~\ref{sec:prelim_findings}), we enforce $\tau_{near} \gg \tau_{hist}$: near-range memory keeps fine-grained tokens, while far-range memory is consolidated into compact episodic slots. Active Forgetting is training-free and introduces no additional trainable parameters. Tunable controls include recent-window size, ratio schedule, and compression operator; implementation details and ablations are analyzed in Sec.~\ref{sec:experiments} and Sec.~\ref{sec:supp_remember_compression_operators}, ~\ref{sec:supp_nearby_window}.

\subsection{Respond (Proactive Response)}
With a compact memory state in hand, the system must next decide \emph{whether} to answer at all. In a continuous stream, queries often arrive before critical visual evidence has appeared; answering prematurely leads to hallucination. Inspired by recent decision-reaction frameworks for real-time interaction~\cite{dispider2025,proactivevideoqa2025}, we place a lightweight readiness gate before expensive reasoning. A readiness head $h$ estimates whether available evidence supports grounded answering:
\begin{equation}
p_{\text{ready}} = h(q_t, M_t).
\end{equation}
Here $p_{\text{ready}}\in[0,1]$ is interpreted as the probability that the current query can be reliably answered by the fast path using the current memory state.
The action rule is threshold-based:
\begin{equation}
a_t=
\begin{cases}
\text{emit }\texttt{<Routine>}, & p_{\text{ready}} < 0.5,\\
\text{continue to Reason}, & \text{otherwise}.
\end{cases}
\end{equation}
For optimization, Respond is trained after Reason SFT cold-start and TB-GRPO are completed: we freeze the fast VLM and train only a lightweight readiness head with SFT labels. Detailed Respond dataset construction (sample collection, labeling, and filtering) is provided in Appendix~\ref{sec:supp_respond}. This decoupled stage keeps additional parameters minimal and avoids perturbing the optimized fast/slow router.

\subsection{Reason (Adaptive Thinking)}
Once a query passes the readiness gate, the system must decide \emph{how much} computation to invest. Not all queries benefit from heavy reasoning---on perception-oriented streaming tasks, a 3B model can match or exceed a 7B thinking model (Sec.~\ref{sec:prelim_findings}, Fig.~\ref{fig:teaser_intro}b). The question is how to identify such queries at decision time without oracle knowledge of their difficulty. We let the fast model either return a direct answer or emit \texttt{<Escalate>} to invoke the slow model. Learning this binary routing policy requires two stages: an SFT cold-start to establish the output format, followed by the proposed TB-GRPO to refine decision quality.

\begin{figure}[t]
  \centering
  \includegraphics[width=\linewidth]{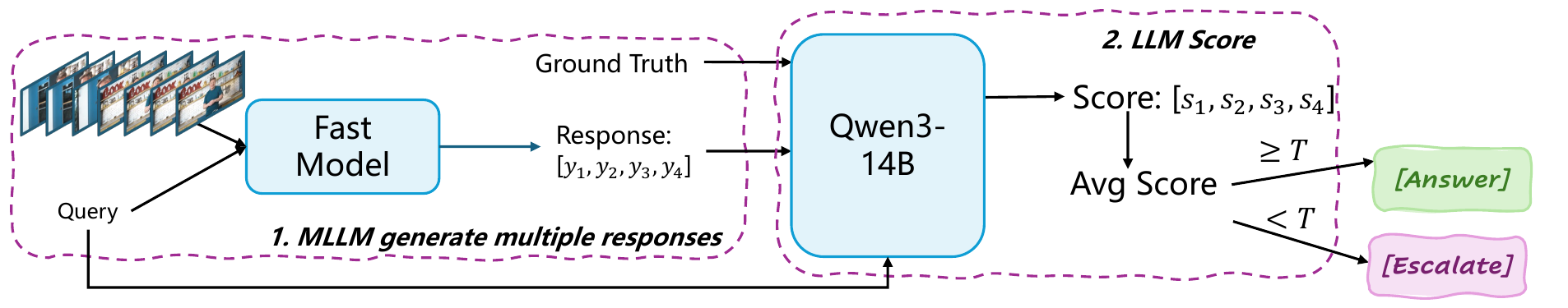}
  \vspace{-6mm}
  \caption{SFT cold-start data pipeline for routing targets. For each query, the fast model samples $K{=}4$ responses. Open-ended responses are scored by an external LLM, and objective responses are checked by exact correctness. The averaged score $\bar{s}$ is thresholded by $T$ to assign \texttt{<Answer>} or \texttt{<Escalate>}. In this figure, $y_i$ denotes the $i$-th sampled response from the fast model.}
  \label{fig:reason_data_pipeline}
\end{figure}

\paragraph{\textbf{Stage 1: SFT cold-start.}}
Before the model can learn \emph{when} to escalate, it must first learn \emph{how}---i.e., reliably produce the routing tokens \texttt{<Answer>} and \texttt{<Escalate>}. We construct answerability supervision via a multi-response pipeline (Fig.~\ref{fig:reason_data_pipeline}). For each training query, the fast model samples $K=4$ responses $\{y^{(k)}\}_{k=1}^{K}$. For open-ended questions, an external LLM assigns quality scores $\{s^{(k)}\}$; for objective questions, we compute binary correctness against ground truth ($s^{(k)}\in\{0,1\}$). We then aggregate
$
\bar{s}=\frac{1}{K}\sum_{k=1}^{K} s^{(k)},
$
and assign routing ground truth by
\begin{equation}
\begin{cases}
\texttt{<Answer>}, & \bar{s}\ge T,\\
\texttt{<Escalate>}, & \bar{s}< T.
\end{cases}
\end{equation}
where $T$ is a quality threshold. The fast model is then fine-tuned on these mixed targets. This stage solves the format problem but not the decision-quality problem: SFT teaches the model to ``speak the routing language'' yet provides no reward signal for making \emph{correct} routing choices.

\paragraph{\textbf{Stage 2: Target-Balanced GRPO (TB-GRPO).}}
The natural next step is RL to refine routing decisions with reward feedback. However, binary routing presents a problem: vanilla GRPO~\cite{deepseekmath2024} rapidly collapses to a single mode---almost always \texttt{<Escalate>}---because the sparse reward signal makes it easier to ``always escalate'' than to discriminate. Prior adaptive-reasoning RL such as AutoThink~\cite{learnwhenthink2025} mitigates early collapse but does not explicitly control escalation frequency, leaving the ratio unpredictable under streaming latency budgets.

\begin{figure*}[t]
  \centering
  \begin{minipage}[t]{0.54\linewidth}
    \centering
    \includegraphics[width=\linewidth]{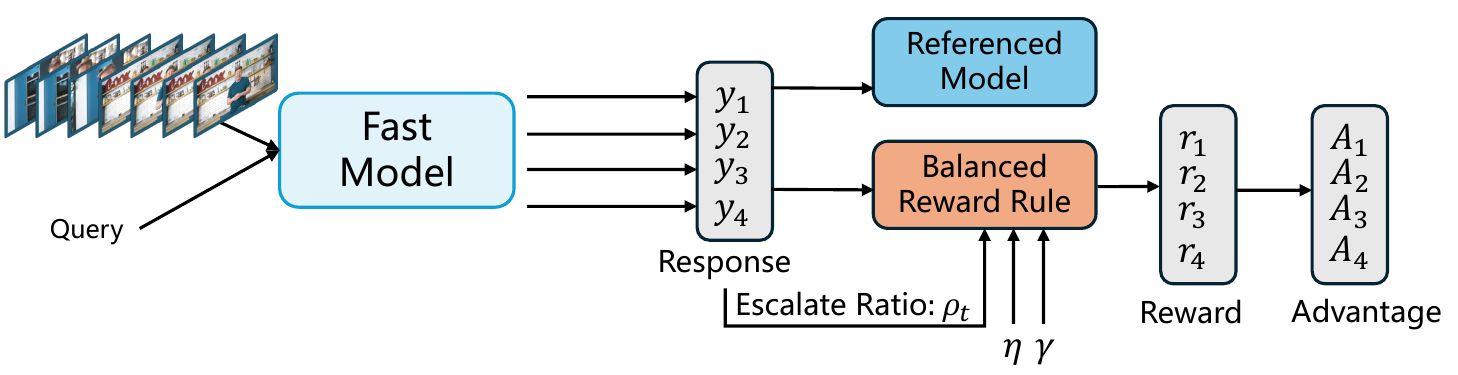}
  \end{minipage}\hfill
  \begin{minipage}[t]{0.43\linewidth}
    \centering
    \includegraphics[width=\linewidth]{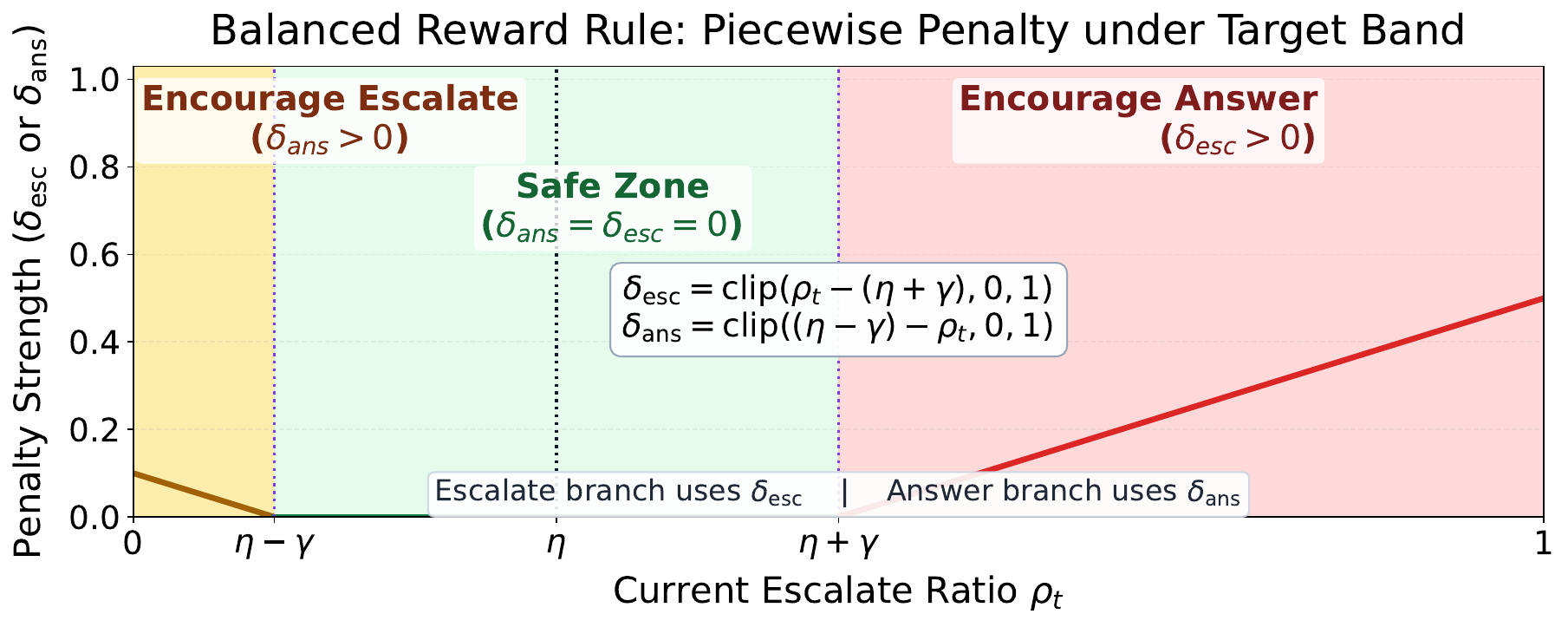}
  \end{minipage}
  \vspace{-3mm}
  \caption{TB-GRPO for adaptive routing. \textbf{Left:} training pipeline where the policy samples grouped routing outputs, computes ratio-aware rewards under target-band control $(\eta,\gamma)$, normalizes advantages, and updates with clipped GRPO plus KL regularization. \textbf{Right:} piecewise penalties versus escalation ratio $\rho$: when $\rho<\eta-\gamma$, non-escalation is penalized ($\delta_{\text{ans}}>0$); when $\eta-\gamma\le\rho\le\eta+\gamma$, both penalties are inactive; when $\rho>\eta+\gamma$, escalation is penalized ($\delta_{\text{esc}}>0$).}
  \label{fig:tbgrpo_framework}
\end{figure*}

TB-GRPO addresses this by introducing \emph{target-band control}: a piecewise penalty mechanism that anchors the escalation ratio around a user-specified operating point $(\eta, \gamma)$, as illustrated in Fig.~\ref{fig:tbgrpo_framework}. Intuitively, when the current escalation ratio $\rho$ drifts above $\eta{+}\gamma$, TB-GRPO suppresses the positive reward for escalation; when $\rho$ drops below $\eta{-}\gamma$, it penalizes non-escalation---effectively applying proportional feedback control to routing frequency.

We now formalize this mechanism. For each query $x$, the fast routing policy $\pi_{\theta_{\text{policy}}}$ samples a response group $\{y_i\}_{i=1}^{G}$, where each $y_i$ is either direct answer text or \texttt{<Escalate>}. Let $e_i=\mathbb{I}[y_i=\texttt{<Escalate>}]$ denote the escalation indicator, $c_i$ the action-dependent correctness, and $\rho=\frac{1}{G}\sum_{i=1}^{G} e_i$ the group escalation ratio.

We first define a base reward that encodes deployment priorities:
\begin{equation}
r_i^{\text{naive}}=
\begin{cases}
2, & e_i=0,\ c_i=1,\\
-1, & e_i=0,\ c_i=0,\\
1, & e_i=1,\ c_i=1,\\
0, & e_i=1,\ c_i=0.
\end{cases}
\end{equation}
A correct direct answer receives a higher reward (2) than a correct escalation (1), because it achieves quality with lower latency on the fast path. This asymmetry encodes the ``answer when capable, escalate when necessary'' principle.

However, $r_i^{\text{naive}}$ alone often leads to early mode collapse. The key idea of TB-GRPO is to modulate rewards based on how far the current escalation ratio $\rho$ deviates from the target band $[\eta{-}\gamma,\ \eta{+}\gamma]$:
\begin{equation}
\begin{aligned}
\delta_{\text{esc}} &=
\mathrm{clip}\!\left(\rho-(\eta+\gamma),0,1\right),\\
\delta_{\text{ans}} &=
\mathrm{clip}\!\left((\eta-\gamma)-\rho,0,1\right).
\end{aligned}
\end{equation}
As Fig.~\ref{fig:tbgrpo_framework} (right) illustrates, $\delta_{\text{esc}}$ activates only when escalation is excessive ($\rho > \eta{+}\gamma$), $\delta_{\text{ans}}$ activates only when escalation is insufficient ($\rho < \eta{-}\gamma$), and both remain zero inside the target band.

These deviations then modulate the base reward:
\begin{equation}
r_i=
\begin{cases}
(1-\delta_{\text{esc}})r_i^{\text{naive}}, & e_i=1,\ c_i=1,\\
(1-\delta_{\text{esc}})r_i^{\text{naive}}-\delta_{\text{esc}}, & e_i=1,\ c_i=0,\\
(1-\delta_{\text{ans}})r_i^{\text{naive}}, & e_i=0,\ c_i=1,\\
(1-\delta_{\text{ans}})r_i^{\text{naive}}-2\delta_{\text{ans}}, & e_i=0,\ c_i=0.
\end{cases}
\end{equation}
The effect is automatic correction: over-escalation compresses escalation rewards, while under-escalation penalizes direct answers, steering $\rho$ back toward the target band.

Finally, group-normalized advantages $A_i=({r_i-\bar{r}})/({{\mathrm{std}(\{r_j\})+\epsilon}})$ are computed and optimized with a clipped policy objective:
\begin{equation}
\begin{aligned}
\mathcal{L}_{\text{TB-GRPO}}
={}&
\mathbb{E}\!\left[
\min\!\left(
w_iA_i,\right.\right.\\
&\left.\left.
\mathrm{clip}(w_i,1-\epsilon_c,1+\epsilon_c)A_i
\right)
\right]\\
&-\beta_{\mathrm{KL}}D_{\mathrm{KL}}(\pi_\theta\|\pi_{\mathrm{ref}}),
\end{aligned}
\end{equation}
where $w_i=\pi_\theta(y_i|x)/\pi_{\theta_{\text{policy}}}(y_i|x)$. The two-parameter control $(\eta, \gamma)$ makes escalation frequency directly tunable for compute budgets.

\begin{figure}[t]
  \centering
  \includegraphics[width=\linewidth]{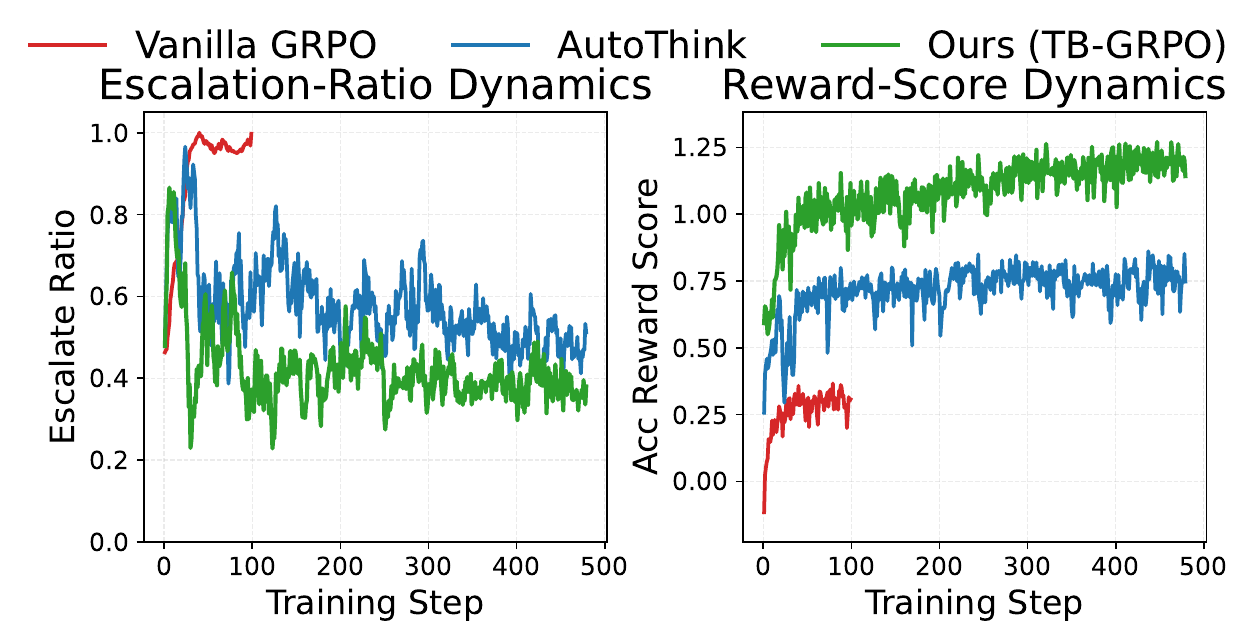}
  \caption{Training dynamics during routing optimization. Vanilla GRPO saturates toward the all-\texttt{<Escalate>} mode, while AutoThink and TB-GRPO avoid collapse. TB-GRPO further maintains a lower escalation ratio and converges to a higher reward level.}
  \label{fig:tbgrpo_training_dynamics}
\end{figure}

Fig.~\ref{fig:tbgrpo_training_dynamics} confirms this empirically. Vanilla GRPO collapses to $\rho{=}1.0$ by step 40. AutoThink~\cite{learnwhenthink2025} avoids early collapse but still settles at a high escalation ratio. TB-GRPO converges to both a lower, more stable ratio and a higher reward level---properties essential for meeting streaming latency constraints. Additional $\eta$ and $\gamma$ ablations appear in Sec.~\ref{sec:experiments} (Table~\ref{tab:ablation_reason_hyperparam}), and the reward-surface visualization is in Appendix~\ref{sec:supp_tbgrpo_reward_surface}.

\section{Experiments}
\label{sec:experiments}


\subsection{Experiment Settings}
\paragraph{Base Settings.}
We use Qwen2.5-VL-3B or Qwen2.5-VL-7B as fast models \cite{qwen25vl2025}. The candidate slow models are Qwen3-VL-4/8B-Thinking \cite{qwen3vl2025}, and Qwen2.5-VL-32B \cite{qwen25vl2025}. In our evaluations, we denote specific pipeline configurations using the format R3-Streaming-[Fast Model]|[Slow Model]. For example, R3-Streaming-3B|4B-Thinking indicates the use of the 3B model for the fast response/routing path and the 4B-Thinking model for the escalated reasoning path. For \textit{Remember}, we adopt the DTD compression operator from TimeChat-Online \cite{timechatonline2025}. Unless otherwise stated, the Nearby threshold is $1.0$ (no compression), the Historical threshold is $0.01$, and the Nearby window is 3 frames. For policy learning, we build the \textit{Respond} and \textit{Reason} training sets from TimeChat-Online-139K \cite{timechatonline2025}. In TB-GRPO, we set the target escalation ratio to $\eta=0.3$ and the tolerance parameter to $\gamma=0.2$.

\paragraph{Benchmarks.}
We evaluate streaming understanding on StreamingBench \cite{streamingbench2024} and OVO-Bench \cite{ovobench2025}, which focus on online video QA. We also conduct long-context evaluation on MLVU \cite{mlvu2024} and Video-MME \cite{videomme2025}. 

\paragraph{Baselines.}
We compare against representative online-video methods, including TimeChat-Online \cite{timechatonline2025}, Streamo \cite{streamo2025}, Dispider \cite{dispider2025}, and StreamingVLM \cite{streamingvlm2025}. We also include Qwen2.5-VL \cite{qwen25vl2025} as a foundation VLM baseline and AdaReTaKe \cite{adaretake2025} as a long-video redundancy-reduction baseline.


\subsection{Streaming Benchmarks}
\begin{table}[t]
\centering
\caption{Main comparison on OVO-Bench \cite{ovobench2025}. We report the average score for each benchmark category and the overall average; detailed subcategory scores are provided in Appendix Table~\ref{tab:supp_main_ovobench_detail}. For our entries, ``$96\%\downarrow$'' denotes the visual-token reduction ratio introduced by \textbf{Remember} (Active Forgetting) before downstream reasoning. Bold indicates the best result within the Streaming MLLMs block.}
\label{tab:main_ovobench}
\scriptsize
\setlength{\tabcolsep}{2.4pt}
\resizebox{\linewidth}{!}{
\begin{tabular}{lcccc}
\toprule
Model & \shortstack{Real-Time\\Avg.} & \shortstack{Backward\\Avg.} & \shortstack{Forward\\Avg.} & Overall Avg. \\
\midrule
\multicolumn{5}{l}{\textbf{Offline Video MLLMs}} \\
Qwen2-VL-72B \cite{qwen25vl2025} & 61.92 & 56.95 & 49.30 & 56.27 \\
Qwen3-VL-4B-Thinking \cite{qwen3vl2025} & 62.93 & 51.56 & 58.74 & 57.74 \\
Qwen2.5-VL-7B \cite{qwen25vl2025} & 64.70 & 44.68 & 41.80 & 50.39 \\
LLaVA-OneVision-7B \cite{llavaonevision2024} & 64.02 & 43.71 & 50.50 & 52.74 \\
Qwen2-VL-7B \cite{qwen25vl2025} & 55.98 & 46.46 & 48.74 & 50.39 \\
LongVU-7B \cite{longvu2024} & 57.61 & 35.01 & 47.50 & 46.71 \\
\midrule
\multicolumn{5}{l}{\textbf{Streaming MLLMs}} \\
Flash-VStream-7B \cite{flashvstream2025} & 28.37 & 27.38 & 45.09 & 33.61 \\
VideoLLM-online-8B \cite{videollmonline2024} & 20.79 & 17.73 & - & - \\
Dispider-7B \cite{dispider2025} & 54.55 & 36.06 & 34.72 & 41.78 \\
ET-Instruct-3B \cite{etinstruct2024} & 46.47 & 28.52 & 46.62 & 40.54 \\
Streamo-3B \cite{streamo2025} & 61.51 & 41.76 & 53.72 & 52.33 \\
Streamo-7B \cite{streamo2025} & 65.98 & 46.10 & 54.77 & 55.61 \\
TimeChat-Online-7B (85\%$\downarrow$) \cite{timechatonline2025} & 58.60 & 42.00 & 36.40 & 45.60 \\
StreamAgent-7B \cite{streamagent2025} & 61.30 & 41.70 & 45.40 & 49.40 \\
FluxMem-7B \cite{fluxmem2026} & - & - & - & 53.40 \\
\rowcolor{oursblue}
R3-Streaming-3B|7B-Instruct (Ours, 96\%$\downarrow$) & 65.12 & 44.64 & 52.44 & 54.07 \\
\rowcolor{oursblue}
R3-Streaming-3B|4B-Thinking (Ours, 96\%$\downarrow$) & 63.58 & 50.23 & \textbf{55.83} & 56.55 \\
\rowcolor{oursblue}
R3-Streaming-7B|4B-Thinking (Ours, 96\%$\downarrow$) & \textbf{71.89} & \textbf{51.27} & 50.60 & \textbf{57.92} \\
\bottomrule
\end{tabular}}
\end{table}

\begin{table}[t]
\centering
\caption{Main comparison on StreamingBench \cite{streamingbench2024}. We report the official ``All'' score in the main paper; detailed subtask scores and frame settings are provided in Appendix Table~\ref{tab:supp_main_streamingbench_detail}. Bold indicates the best result within the Streaming MLLMs block.}
\label{tab:main_streamingbench}
\scriptsize
\setlength{\tabcolsep}{4pt}
\resizebox{\linewidth}{!}{
\begin{tabular}{lcc}
\toprule
Model & Params & All \\
\midrule
\multicolumn{3}{l}{\textbf{Human}} \\
Human & - & 91.46 \\
\midrule
\multicolumn{3}{l}{\textbf{Proprietary MLLMs}} \\
GPT-4o \cite{gpt4o2024} & - & 73.28 \\
Claude 3.5 Sonnet \cite{claude35sonnet2024} & - & 72.44 \\
\midrule
\multicolumn{3}{l}{\textbf{Offline Video MLLMs}} \\
LLaVA-OneVision \cite{llavaonevision2024} & 7B & 71.12 \\
Qwen2-VL \cite{qwen25vl2025} & 7B & 69.04 \\
Qwen3-VL-4B-Thinking \cite{qwen3vl2025} & 4B & 73.16 \\
Qwen2.5-VL-7B \cite{qwen25vl2025} & 7B & 73.28 \\
MiniCPM-V 2.6 \cite{minicpmv26_2024} & 8B & 67.44 \\
LLaVA-NeXT-Video \cite{llavavideo2024} & 32B & 66.96 \\
VILA-1.5 \cite{vila2023} & 8B & 52.32 \\
Video-CCAM \cite{videoccam2024} & 14B & 53.96 \\
Video-LLaMA2 \cite{videollama22024} & 7B & 49.52 \\
\midrule
\multicolumn{3}{l}{\textbf{Streaming MLLMs}} \\
Flash-VStream \cite{flashvstream2025} & 7B & 23.23 \\
VideoLLM-online \cite{videollmonline2024} & 8B & 35.99 \\
Dispider \cite{dispider2025} & 7B & 67.63 \\
TimeChat-Online-7B (83\%$\downarrow$) \cite{timechatonline2025} & 7B & 73.64 \\
StreamAgent \cite{streamagent2025} & 7B & 74.28 \\
\rowcolor{oursblue}
R3-Streaming-3B|7B-Instruct (Ours, 95\%$\downarrow$) & 3B/7B & 73.84 \\
\rowcolor{oursblue}
R3-Streaming-3B|4B-Thinking (Ours, 95\%$\downarrow$) & 3B/4B & 74.36 \\
\rowcolor{oursblue}
R3-Streaming-7B|4B-Thinking (Ours, 95\%$\downarrow$) & 7B/4B & \textbf{76.36} \\
\bottomrule
\end{tabular}}
\end{table}

As shown in Table~\ref{tab:main_ovobench} and Table~\ref{tab:main_streamingbench}, R3-Streaming achieves SOTA performance among streaming MLLMs on two mainstream streaming benchmarks, OVO-Bench~\cite{ovobench2025} and StreamingBench~\cite{streamingbench2024}, outperforming recent streaming methods. While reducing more than 95\% of visual tokens, our method also surpasses both the slow baseline Qwen3-VL-4B-Thinking and the fast baseline Qwen2.5-VL-7B. This indicates that the performance gain does not simply come from a fixed trade-off between slow and fast models. Instead, R3-Streaming uses agentic control to adaptively select a more suitable action according to the query type. The finer-grained subtask analysis in Fig.~\ref{fig:supp_streamingbench_subtasks} further corroborates this behavior.

\subsection{Long Video Understanding Benchmarks}
\begin{table}[t]
\centering
\caption{Long-video understanding results on MLVU \cite{mlvu2024} and Video-MME \cite{videomme2025}. For Video-MME, we report the overall score without subtitles. Bold indicates the best value per metric.}
\label{tab:main_long_video}
\footnotesize
\setlength{\tabcolsep}{4pt}
\resizebox{\linewidth}{!}{
\begin{tabular}{lcc}
\toprule
Model & MLVU & \shortstack{Video-MME\\Overall} \\
\midrule
Baseline (Qwen2.5-VL-3B \cite{qwen25vl2025}) & 66.5 & 60.4 \\
Qwen2.5-VL-7B \cite{qwen25vl2025} & 66.9 & 63.6 \\
VideoAgent \cite{videoagent2024} & 57.8 & 56.0 \\
TimeChat-Online-7B \cite{timechatonline2025} & 62.9 & 63.3 \\
StreamAgent-7B \cite{streamagent2025} & 67.2 & 62.9 \\
AdaReTaKe \cite{adaretake2025} & - & 63.1 \\
\midrule
R3-Streaming (Ours, 3B|4B) & \textbf{70.6} & \textbf{65.5} \\
\bottomrule
\end{tabular}
}
\end{table}

Though designed for streaming, R3-Streaming generalizes to offline long-video tasks (Table~\ref{tab:main_long_video}). Considering the substantial differences between offline and online video settings, we reset the Historical threshold to 0.5 for offline long-video evaluation, corresponding to about 45\% token drop. A more detailed hyperparameter analysis is provided in Appendix~\ref{sec:supp_backbone_grids} (Fig.~\ref{fig:supp_backbone_grids}) and Sec.~\ref{sec:supp_offline_long_video}.
This hyperparameter difference reflects the offline/online setting shift, and we use the same offline configuration across long-video benchmarks.
On both benchmarks, R3-Streaming also outperforms representative long-video methods such as AdaReTaKe~\cite{adaretake2025} and VideoAgent~\cite{videoagent2024}. These results confirm the robustness of our method: although R3-Streaming is designed for online/streaming scenarios, it still achieves strong performance on offline scenarios.

\subsection{Ablations}

\begin{table*}[t]
\centering
\begingroup
\setlength{\tabcolsep}{4pt}
\renewcommand{\arraystretch}{1.08}
\newsavebox{\abloneA}
\newsavebox{\abloneB}
\newsavebox{\abloneC}
\newlength{\abloneH}
\sbox{\abloneA}{%
  \begin{minipage}[t]{0.325\textwidth}
  \centering
  \captionof{table}{Base-module ablation on StreamingBench \cite{streamingbench2024} and OVO-Bench \cite{ovobench2025}. Integrating both \textbf{Remember} and \textbf{Reason} yields the best performance.}
  \label{tab:ablation_base_module}
  \footnotesize
  \resizebox{\linewidth}{!}{
  \begin{tabular}{lcc}
  \toprule
  Method & \shortstack{Streaming-\\Bench $\uparrow$} & \shortstack{OVO-\\Bench $\uparrow$} \\
  \midrule
  Baseline (Qwen2.5-VL-3B \cite{qwen25vl2025}) & 68.6 & 51.4 \\
  w/ Remember & 71.0 & 52.8 \\
  w/ Reason & 73.4 & 55.9 \\
  w/ Both & \textbf{74.4} & \textbf{56.6} \\
  \bottomrule
  \end{tabular}
  }
  \end{minipage}%
}
\sbox{\abloneB}{%
  \begin{minipage}[t]{0.325\textwidth}
  \centering
  \captionof{table}{Remember compression comparison on Qwen2.5-VL-7B \cite{qwen25vl2025}. Our lightweight DTD-based variant achieves the best accuracy and highest drop ratio.}
  \vspace{-2mm}
  \label{tab:ablation_compression_method}
  \footnotesize
  \resizebox{\linewidth}{!}{
  \begin{tabular}{lcc}
  \toprule
  Method & StreamingBench $\uparrow$ & Drop Ratio $\uparrow$ \\
  \midrule
  DivPrune \cite{divprune2025} & 68.48 & 92.8\% \\
  VisionZip \cite{visionzip2025} & 68.32 & 92.8\% \\
  DTD \cite{timechatonline2025} & 65.16 & 92.7\% \\
  Ours w/ DTD & \textbf{75.90} & \textbf{95.0\%} \\
  \bottomrule
  \end{tabular}
  }
  \end{minipage}%
}
\sbox{\abloneC}{%
  \begin{minipage}[t]{0.325\textwidth}
  \centering
  \vspace{1mm}
  \captionof{table}{Respond ablation on the StreamingBench \cite{streamingbench2024} Proactive split. The dedicated readiness head substantially improves proactive output quality.}
  \label{tab:ablation_respond}
  \footnotesize
  \resizebox{\linewidth}{!}{
  \begin{tabular}{lc}
  \toprule
  Method & Proactive Output $\uparrow$ \\
  \midrule
  Qwen2.5-VL-7B \cite{qwen25vl2025} & 0.204 \\
  Qwen2.5-VL-3B \cite{qwen25vl2025} & 0.216 \\
  Ours & \textbf{0.328} \\
  \bottomrule
  \end{tabular}
  }
  \end{minipage}%
}
\setlength{\abloneH}{\dimexpr\ht\abloneA+\dp\abloneA\relax}
\ifdim\dimexpr\ht\abloneB+\dp\abloneB\relax>\abloneH
  \setlength{\abloneH}{\dimexpr\ht\abloneB+\dp\abloneB\relax}
\fi
\ifdim\dimexpr\ht\abloneC+\dp\abloneC\relax>\abloneH
  \setlength{\abloneH}{\dimexpr\ht\abloneC+\dp\abloneC\relax}
\fi
\begin{minipage}[t][\abloneH][t]{0.325\textwidth}\usebox{\abloneA}\vfill\end{minipage}\hfill
\begin{minipage}[t][\abloneH][t]{0.325\textwidth}\usebox{\abloneB}\vfill\end{minipage}\hfill
\begin{minipage}[t][\abloneH][t]{0.325\textwidth}\usebox{\abloneC}\vfill\end{minipage}
\endgroup
\end{table*}

\begin{table*}[t]
\centering
\begingroup
\setlength{\tabcolsep}{4pt}
\renewcommand{\arraystretch}{1.08}
\newsavebox{\abltwoA}
\newsavebox{\abltwoB}
\newsavebox{\abltwoC}
\newlength{\abltwoH}
\sbox{\abltwoA}{%
  \begin{minipage}[t]{0.285\textwidth}
  \centering
  \captionof{table}{Comparison of Reason routing policies. TB-GRPO avoids collapse and achieves the strongest performance-efficiency trade-off.}
  \vspace{-2mm}
  \label{tab:ablation_reason_method}
  \footnotesize
  \resizebox{\linewidth}{!}{
  \begin{tabular}{lcc}
  \toprule
  Method & \shortstack[c]{Streaming-\\Bench $\uparrow$} & \shortstack[c]{Escalate\\Ratio $\downarrow$} \\
  \midrule
  SFT-only & 73.16 & 100.0\% \\
  Vanilla GRPO \cite{deepseekmath2024} & 73.16 & 100.0\% \\
  AutoThink \cite{learnwhenthink2025} & 70.00 & 53.2\% \\
  Ours (TB-GRPO) & \textbf{74.36} & \textbf{24.0\%} \\
  \bottomrule
  \end{tabular}
  }
  \end{minipage}%
}
\sbox{\abltwoB}{%
  \begin{minipage}[t]{0.285\textwidth}
  \centering
  \captionof{table}{TB-GRPO hyperparameter ablation (\textit{Acc. / Esc. ratio}). $(\eta,\gamma)$ controls the operating point between accuracy and escalation frequency.}
  \label{tab:ablation_reason_hyperparam}
  \footnotesize
  \resizebox{\linewidth}{!}{
  \begin{tabular}{c|ccc}
  \toprule
  $\eta \backslash \gamma$ & 0.1 & 0.2 & 0.3 \\
  \midrule
  0.2 & 71.1 / 15\% & 73.2 / 19\% & 74.2 / 26\% \\
  0.3 & 73.8 / 21\% & 74.4 / 24\% & 74.1 / 32\% \\
  0.4 & 74.5 / 32\% & 73.5 / 35\% & 73.7 / 41\% \\
  \bottomrule
  \end{tabular}
  }
  \end{minipage}%
}
\sbox{\abltwoC}{%
  \begin{minipage}[t]{0.40\textwidth}
  \centering
  \captionof{figure}{Efficiency vs Performance. Adaptive routing consistently outperforms direct slow-only inference across all tested slow models.}
  \vspace{-4mm}
  \label{fig:ablation_reason_model}
  \includegraphics[width=\linewidth]{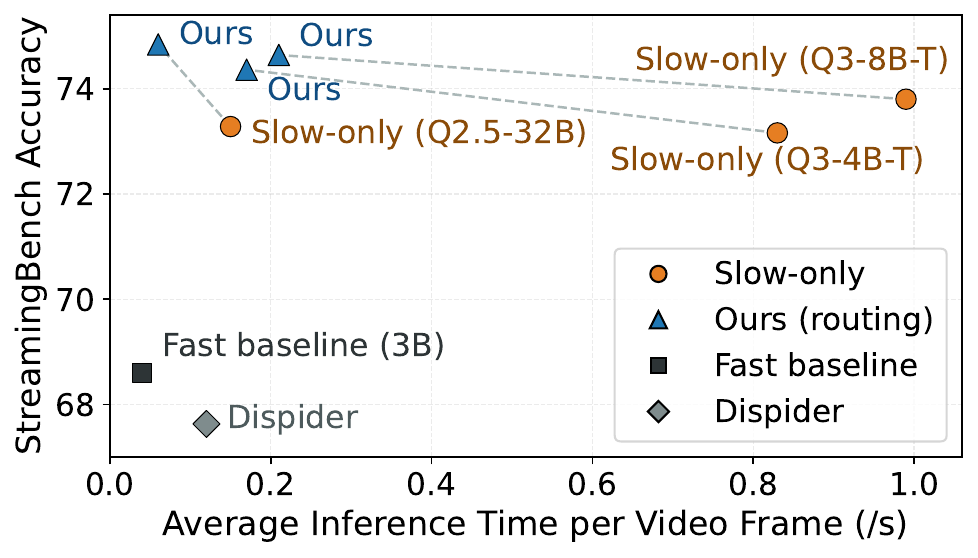}
  \end{minipage}%
}
\setlength{\abltwoH}{\dimexpr\ht\abltwoA+\dp\abltwoA\relax}
\ifdim\dimexpr\ht\abltwoB+\dp\abltwoB\relax>\abltwoH
  \setlength{\abltwoH}{\dimexpr\ht\abltwoB+\dp\abltwoB\relax}
\fi
\ifdim\dimexpr\ht\abltwoC+\dp\abltwoC\relax>\abltwoH
  \setlength{\abltwoH}{\dimexpr\ht\abltwoC+\dp\abltwoC\relax}
\fi
\begin{minipage}[t][\abltwoH][t]{0.285\textwidth}\usebox{\abltwoA}\vfill\end{minipage}\hfill
\begin{minipage}[t][\abltwoH][t]{0.285\textwidth}\usebox{\abltwoB}\vfill\end{minipage}\hfill
\begin{minipage}[t][\abltwoH][t]{0.40\textwidth}\usebox{\abltwoC}\vfill\end{minipage}
\endgroup
\end{table*}

\subsubsection{Base Module Ablation}
We first evaluate the individual and combined effects of the two main technical components, Remember and Reason, in Table~\ref{tab:ablation_base_module}. Respond only targets proactive forward tasks rather than all tasks accuracy, and is therefore evaluated separately on the StreamingBench Proactive split in Sec.~\ref{sec:respond_ablation}. Applying either the Remember (memory compression) or Reason (adaptive routing) module independently improves upon the Qwen2.5-VL-3B baseline \cite{qwen25vl2025}. Crucially, integrating both modules yields the best performance on StreamingBench \cite{streamingbench2024} and OVO-Bench \cite{ovobench2025}. This demonstrates that active memory control and adaptive compute routing are not isolated optimizations, but rather highly complementary components within our agentic framework.

\subsubsection{Remember Compression Ablation}
\label{sec:compression_ablation}

In Table~\ref{tab:ablation_compression_method}, we compare R3-Streaming with prior methods that perform token compression alone. The results show that age-aware policy brings a substantial performance gain over standalone token compression. Furthermore, Fig.~\ref{fig:ablation_remember_methods} in Appendix~\ref{sec:supp_remember_compression_operators} studies the effect of different compression operators. We find that different operators achieve similar peak performance under the age-aware policy, indicating that the gain comes more from the age-aware policy than from the specific operator design. This also demonstrates the robustness of our age-aware memory strategy. Appendix~\ref{app:memory route} provides a finer-grained view of this complementarity, showing that better nearby-memory preservation simultaneously improves accuracy and reduces the escalation ratio.

\subsubsection{Respond Proactive Ablation}
\label{sec:respond_ablation}
Table~\ref{tab:ablation_respond} shows that the Respond head improves proactive output performance on StreamingBench \cite{streamingbench2024} (0.328 vs.\ 0.216 with the same fast backbone). This suggests that readiness control is learned by the response head instead of coming from base-model scaling.



\subsubsection{Reason Ablation}
\label{sec:model_size_analysis}
In Table~\ref{tab:ablation_reason_method}, we isolate the effect of the proposed TB-GRPO objective. Directly optimizing agentic control with Vanilla GRPO~\cite{deepseekmath2024} or SFT-only leads to mode collapse, where all queries trigger \texttt{<Escalate>} (invoking the slow model), causing large latency as shown in Fig.~\ref{fig:ablation_reason_model}. Also, AutoThink~\cite{learnwhenthink2025} still invokes escalation for a large fraction of queries, while the performance is suboptimal. By contrast, TB-GRPO triggers only a small number of escalations, improving benchmark performance while maintaining low latency, making it better suited to streaming understanding. Appendix~\ref{app:task_route} further shows that this routing policy accurately adapts to intrinsic task difficulty.

\subsection{Complexity and Cost Analysis}
\label{sec:complexity_cost}
In Fig.~\ref{fig:ablation_reason_model}, we analyze the complexity of R3-Streaming. Compared with the fast baseline and the prior streaming method Dispider~\cite{dispider2025}, our method substantially improves benchmark performance. Compared with slow-only models, R3-Streaming greatly reduces per-frame latency. This shows that our method can preserve strong performance while remaining practical for real-time streaming scenarios.


\section{Conclusion}
We present R3-Streaming, a cascaded agentic framework that decomposes streaming video understanding into three sequential decisions, enabling adaptive coordination of memory compression, response timing, and compute routing. Stabilized by TB-GRPO, our system achieves state-of-the-art streaming performance on OVO-Bench and StreamingBench while reducing visual token usage by 95–96\%, and generalizes effectively to long-video scenarios.

\section*{Limitations}
While TB-GRPO successfully constrains the average escalation ratio to respect overarching compute budgets, handling worst-case latency spikes during prolonged segments of extreme information density remains an open challenge. Current streaming benchmarks primarily evaluate average metrics. We believe that addressing such localized computational peaks, where even fast models struggle, is an important next step for the streaming VLM community and calls for spike-aware benchmarks in future work.

\bibliography{main}

\clearpage
\appendix

\section{Appendix Overview and Benchmark Task Taxonomy}
\label{sec:supp_overview}

This appendix provides supplementary analyses and full results organized by the three core modules of R3-Streaming.
Appendix~\ref{sec:supp_remember} consolidates all \textbf{Remember} (Active Forgetting) experiments: compression operator ablation (Sec.~\ref{sec:supp_remember_compression_operators}), backbone and benchmark generalization grids (Sec.~\ref{sec:supp_backbone_grids}), and nearby window-size analysis (Sec.~\ref{sec:supp_nearby_window}).
Appendix~\ref{sec:supp_respond} describes the \textbf{Respond} module's SFT data construction.
Appendix~\ref{sec:supp_reason} presents additional \textbf{Reason} (Adaptive Thinking) analysis: the TB-GRPO reward-surface visualization (Sec.~\ref{sec:supp_tbgrpo_reward_surface}) and subtask-level routing behavior.
Finally, Appendix~\ref{sec:supp_full_results} collects the full subtask-level benchmark tables.
We begin by spelling out the subtask taxonomies used throughout our evaluation tables so that the abbreviated column headers remain easy to read.

\paragraph{OVO-Bench.}
OVO-Bench~\cite{ovobench2025} evaluates online video understanding under three causal settings: \textit{Real-Time Visual Perception}, \textit{Backward Tracing}, and \textit{Forward Active Responding}. Table~\ref{tab:supp_ovobench_taxonomy} lists the fine-grained task names. The official project page uses \texttt{FTP} for \emph{Future Prediction}, while several reproduced result tables, including ours, inherit the shorthand \texttt{FPD}. We keep \texttt{FPD} in the tables for consistency with the main paper and interpret it as \emph{Future Prediction}.

\begin{table}[ht]
\centering
\caption{\textbf{OVO-Bench Subtask Taxonomy.} Detailed full names for the subtask abbreviations referenced in the main paper's evaluation tables.}
\label{tab:supp_ovobench_taxonomy}
\small
\setlength{\tabcolsep}{4pt}
\resizebox{\linewidth}{!}{%
\begin{tabular}{lll}
\toprule
Group & Abbrev. & Full Name \\
\midrule
Real-Time Visual Perception & OCR & Optical Character Recognition \\
Real-Time Visual Perception & ACR & Action Recognition \\
Real-Time Visual Perception & ATR & Attribute Recognition \\
Real-Time Visual Perception & STU & Spatial Understanding \\
Real-Time Visual Perception & FPD & Future Prediction \\
Real-Time Visual Perception & OJR & Object Recognition \\
Backward Tracing & EPM & Episodic Memory \\
Backward Tracing & ASI & Action Sequence Identification \\
Backward Tracing & HLD & Hallucination Detection \\
Forward Active Responding & REC & Repetition Event Count \\
Forward Active Responding & SSR & Sequential Steps Recognition \\
Forward Active Responding & CRR & Clues Reveal Responding \\
\bottomrule
\end{tabular}
}
\end{table}

\paragraph{StreamingBench.}
In the main paper we report the \emph{Real-Time Visual Understanding} split of StreamingBench~\cite{streamingbench2024}. Table~\ref{tab:supp_streamingbench_taxonomy} expands the ten abbreviated subtask names used in StreamingBench.

\begin{table}[ht]
\centering
\caption{\textbf{StreamingBench Subtask Taxonomy.} Detailed full names for the real-time visual understanding subtasks referenced in the main paper.}
\label{tab:supp_streamingbench_taxonomy}
\small
\setlength{\tabcolsep}{4pt}
\begin{tabular}{ll}
\toprule
Abbrev. & Full Name \\
\midrule
OP & Object Perception \\
CR & Causal Reasoning \\
CS & Clip Summarization \\
ATP & Attribute Perception \\
EU & Event Understanding \\
TR & Text-Rich Understanding \\
PR & Prospective Reasoning \\
SU & Spatial Understanding \\
ACP & Action Perception \\
CT & Counting \\
\bottomrule
\end{tabular}
\end{table}




\section{Remember: Memory Compression Details}
\label{sec:supp_remember}

This section consolidates all ablation and generalization experiments related to the Remember (Active Forgetting) module.

\subsection{Compression Operator Ablation}
\label{sec:supp_remember_compression_operators}
\begin{figure*}[ht]
  \centering
  \includegraphics[width=\linewidth]{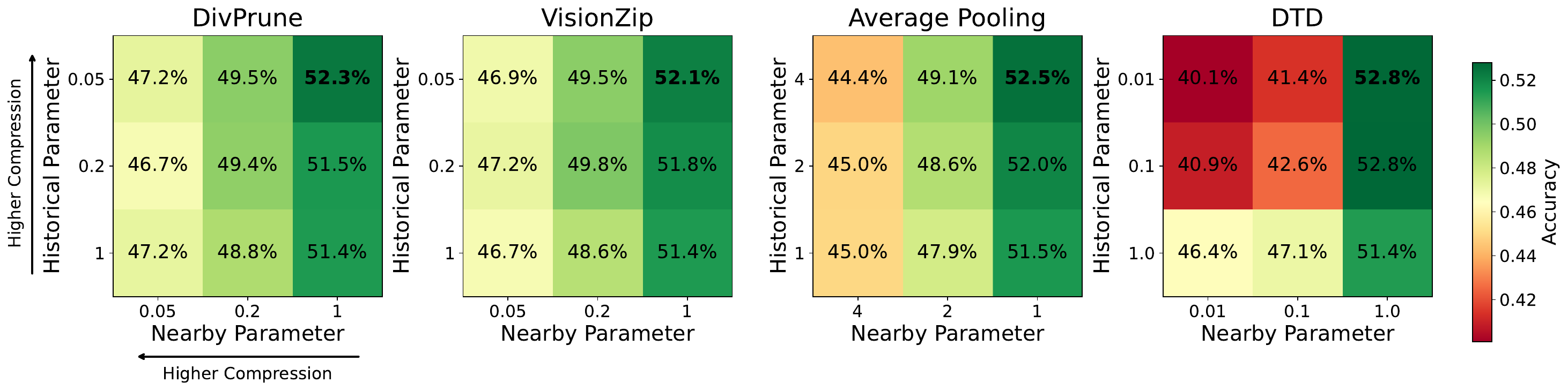}
  \caption{\textbf{Remember ablation with four compression operators on OVO-Bench \cite{ovobench2025}.} Each panel shows a grid search over operator-specific hyperparameters, and each cell reports overall accuracy. For Pooling, Parameter indicates the pooling kernel size. In the top-right region (aggressive historical compression with nearby evidence preserved), all operators outperform the no-compression baseline.}
  \label{fig:ablation_remember_methods}
\end{figure*}

Figure~\ref{fig:ablation_remember_methods} shows that Remember does not rely on a single compression operator. Across DivPrune~\cite{divprune2025}, VisionZip~\cite{visionzip2025}, Average Pooling, and DTD~\cite{timechatonline2025}, the strongest cells consistently appear when nearby evidence is preserved and historical memory is aggressively compressed. This supports the main-paper conclusion that the age-aware compression policy matters more than the particular operator used to instantiate it.

\subsection{Compression Grids Across Backbones and Benchmarks}
\label{sec:supp_backbone_grids}
\begin{figure*}[ht]
  \centering
  \includegraphics[width=\linewidth]{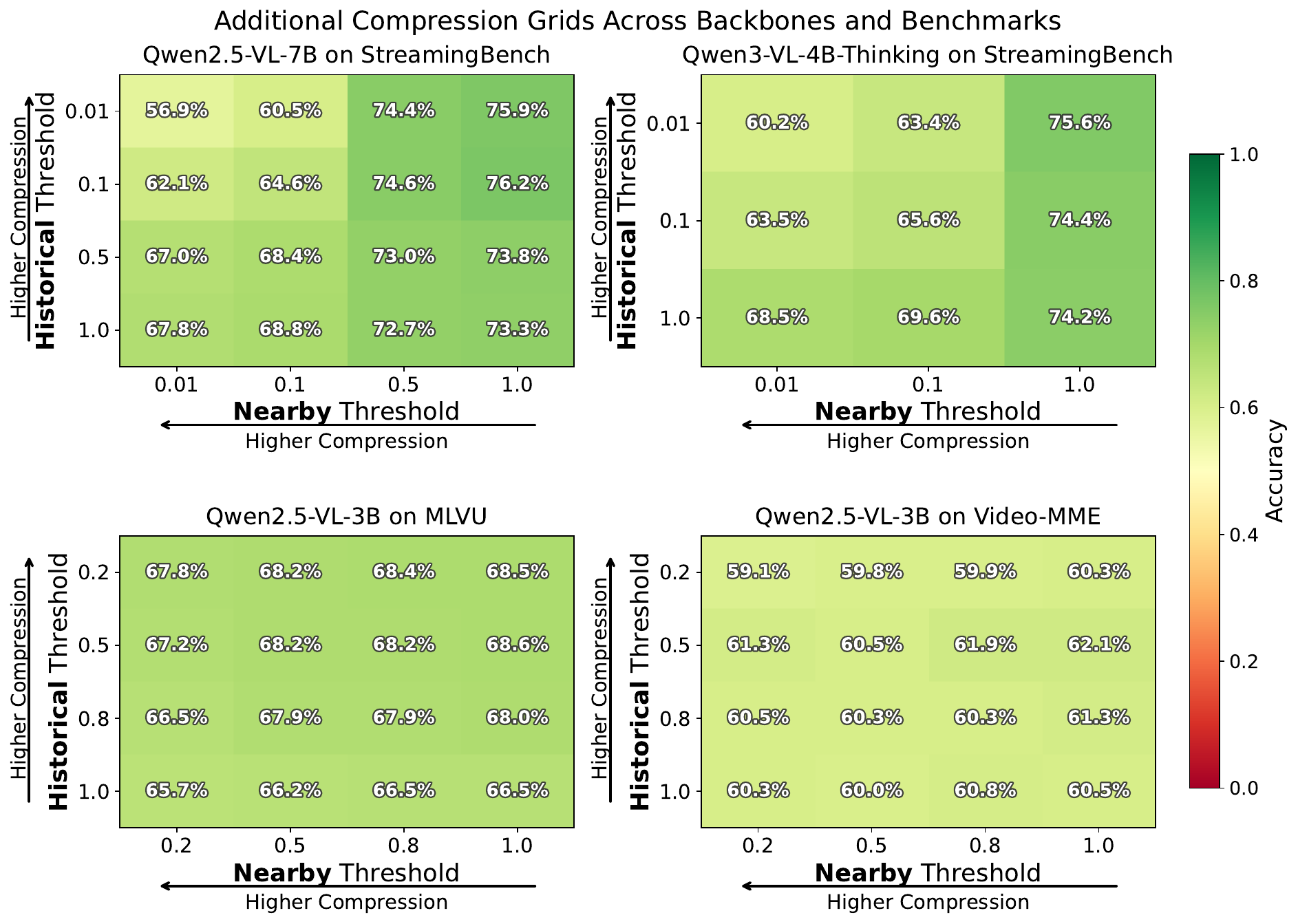}
  \caption{\textbf{Additional memory compression grids across backbones and benchmarks.} The heatmaps illustrate the effect of varying the Historical and Nearby thresholds on overall accuracy. For streaming tasks (top row), the optimal operating region consistently lies in the top-right (Historical=0.01, Nearby=1.0), validating that our recent-focused \textit{Active Forgetting} policy is universally effective across both fast models (Qwen2.5-VL-7B) and reasoning models (Qwen3-VL-4B-Thinking). For offline long-video tasks (bottom row), we expand the sweep with threshold 0.2; the core robustness remains, but moderate historical compression is preferred because useful evidence is more evenly distributed across the video.}
  \label{fig:supp_backbone_grids}
\end{figure*}

\subsubsection{Consistency Across Model Architectures}
For streaming video understanding (StreamingBench), the compression heatmaps reveal a highly consistent pattern regardless of the underlying model. Whether using a standard fast model (Qwen2.5-VL-7B) or a reasoning-heavy model (Qwen3-VL-4B-Thinking), the optimal performance invariably lies in the top-right region of the grid. Specifically, maintaining recent evidence at high fidelity (Nearby Threshold = 1.0) while aggressively compressing stale history (Historical Threshold = 0.01) yields the highest accuracy: 75.9\% for the 7B model and 75.6\% for the 4B-Thinking model. Moving horizontally to the left (compressing nearby frames) causes a catastrophic performance drop---for instance, dropping from 75.9\% to 56.9\% on the 7B model at Historical = 0.01. Conversely, moving vertically downward (retaining more history) provides no meaningful accuracy gain and often degrades performance by introducing noise. This definitively proves that the age-aware Active Forgetting policy is not backbone-specific; the optimal strategy is universally recent-focused.

\subsubsection{Generalization to Offline Long-Video Tasks}
\label{sec:supp_offline_long_video}
The bottom panels of Figure~\ref{fig:supp_backbone_grids} extend this analysis to offline long-video benchmarks (MLVU~\cite{mlvu2024} and Video-MME~\cite{videomme2025}) using the Qwen2.5-VL-3B backbone and an expanded threshold sweep $\{0.2,0.5,0.8,1.0\}$. Although our method is designed for streaming, it remains robust in offline long-video settings. Because offline videos distribute useful evidence more evenly across the full clip, the best operating point shifts from extreme historical compression to a moderate historical threshold (Historical Threshold = 0.5): 68.6\% on MLVU and 62.1\% on Video-MME when Nearby Threshold = 1.0. This adjustment is not a video-specific hyperparameter search, but a deterministic scenario toggle: in deployment, the system knows a priori whether it is processing a continuous live stream or a pre-recorded offline file, so this binary configuration adapts R3 to the distinct evidence distributions of online and offline settings without requiring complex adaptive tuning. The added 0.2 threshold confirms this distinction: aggressive historical compression remains competitive on MLVU (68.5\% at Historical = 0.2, Nearby = 1.0) but is less reliable on Video-MME (60.3\%), where retaining moderate history better supports long-range dependency modeling. Taken together, these grids show that our dual-zone memory policy robustly transfers beyond streaming while allowing the historical threshold to be selected by the known deployment scenario.

\subsection{Effect of Nearby Window Size}
\label{sec:supp_nearby_window}
\begin{table}[ht]
\centering
\caption{\textbf{Effect of Nearby window size on StreamingBench performance and compression ratio.} A window of 3 frames rigorously enforces the extreme 95\% token drop ratio utilized as the default in our main experiments. While expanding the window to 5--10 frames yields peak accuracy, excessive expansion (e.g., 20 or 40 frames) significantly degrades performance despite retaining more tokens. This confirms that \textit{Active Forgetting} crucially filters out informational noise from stale history.}
\label{tab:supp_nearby_window}
\small
\setlength{\tabcolsep}{6pt}
\resizebox{\linewidth}{!}{%
\begin{tabular}{ccc}
\toprule
Nearby Window & StreamingBench Score $\uparrow$ & Drop Ratio $\uparrow$ \\
\midrule
3 & 71.0 & 95\% \\
5 & 72.24 & 92\% \\
10 & 72.08 & 87\% \\
20 & 70.40 & 77\% \\
40 & 69.88 & 56\% \\
\bottomrule
\end{tabular}
}
\end{table}

Table~\ref{tab:supp_nearby_window} quantifies the performance-efficiency trade-off controlled by the Nearby window size.

\subsubsection{Balancing Compression and Accuracy}
The size of the Nearby window serves as a flexible tuning knob depending on deployment constraints. A strict window of 3 frames achieves an extreme 95\% token drop ratio, making it highly effective for strictly constrained streaming environments. However, expanding the window to 5 or 10 frames unlocks the highest accuracy regime (peaking at 72.24) while still maintaining a highly efficient 87\%-92\% drop ratio. This demonstrates that our framework easily adapts to scenarios where a slight relaxation of compute budgets can be traded for peak performance.

\subsubsection{The Window Size of Active Forgetting}
Most importantly, Table~\ref{tab:supp_nearby_window} shows that indiscriminately increasing the window size eventually harms performance. When expanded to 40 frames, the accuracy drops to 69.88, even though the model retains significantly more visual tokens (only a 56\% drop ratio). This counter-intuitive degradation directly corroborates Finding 1 from the main text: preserving too much uncompressed history introduces informational noise that dilutes critical recent evidence. Thus, proper window sizing in Active Forgetting acts not merely as a latency-saving heuristic, but as an essential noise-filtering mechanism that actively improves model perception.

\section{Respond: Proactive Response SFT Data Construction}
\label{sec:supp_respond}

As introduced in the main text, the \textit{Respond} module uses a lightweight readiness head to emit \texttt{<Routine>} when evidence is insufficient, deferring the answer until a reliable response can be grounded. We construct the training set for this proactive behavior using streaming video data from TimeChat-Online-139K~\cite{timechatonline2025} and COIN~\cite{tang2019coin}.

\noindent\textbf{Decision-Boundary-Focused Sampling.}
Rather than randomly sampling positive and negative frames across an entire video, we focus exclusively on temporal segments where answerability strictly hinges on the arrival of a specific visual clue. For each training instance, we identify the exact clue timestamp $t_c$ at which the question first becomes answerable. We then convert the immediate temporal neighborhood into binary readiness supervision:

\begin{itemize}
    \item \textbf{Insufficient Evidence (Target: \texttt{<Routine>}):} The three frames immediately preceding the clue emergence, namely $\{t_c - 3, t_c - 2, t_c - 1\}$, are explicitly labeled as unready.
    \item \textbf{Ready (Target: Proceed to \textit{Reason}):} The frame at the clue timestamp and the subsequent two frames, namely $\{t_c, t_c + 1, t_c + 2\}$, are labeled as ready.
\end{itemize}

Boundary cases are clipped to the valid range of the video, and frames outside this narrow local supervision window are deliberately omitted. This hard-mining approach ensures the supervision remains heavily concentrated on the exact decision boundary. By forcing the readiness head to distinguish between the frame \textit{just before} and \textit{just after} the evidence appears, the model learns to avoid premature hallucination while minimizing reaction latency, satisfying the strict causal constraints of streaming video understanding. Finally, as described in the main paper, we freeze the fast VLM backbone and train only the lightweight readiness head using this targeted binary data.

\section{Reason: Adaptive Thinking (TB-GRPO) Details}
\label{sec:supp_reason}

This section presents additional analysis of the Reason module, including the TB-GRPO reward-surface visualization and subtask-level routing behavior.
\subsection{SFT Data Construction}
As shown in Fig.~\ref{fig:reason_data_pipeline}, we construct a portion of data for the cold start prior to TB-GRPO. Specifically for the LLM Score, we employ an LLM to compare the generated responses with the ground truth (GT) and assign scores using a 5-point rating scale. The threshold $T$ is empirically set to 2.5.

\subsection{Reward-Surface Visualization}
\label{sec:supp_tbgrpo_reward_surface}

\begin{figure*}[t]
  \centering
  \includegraphics[width=\linewidth]{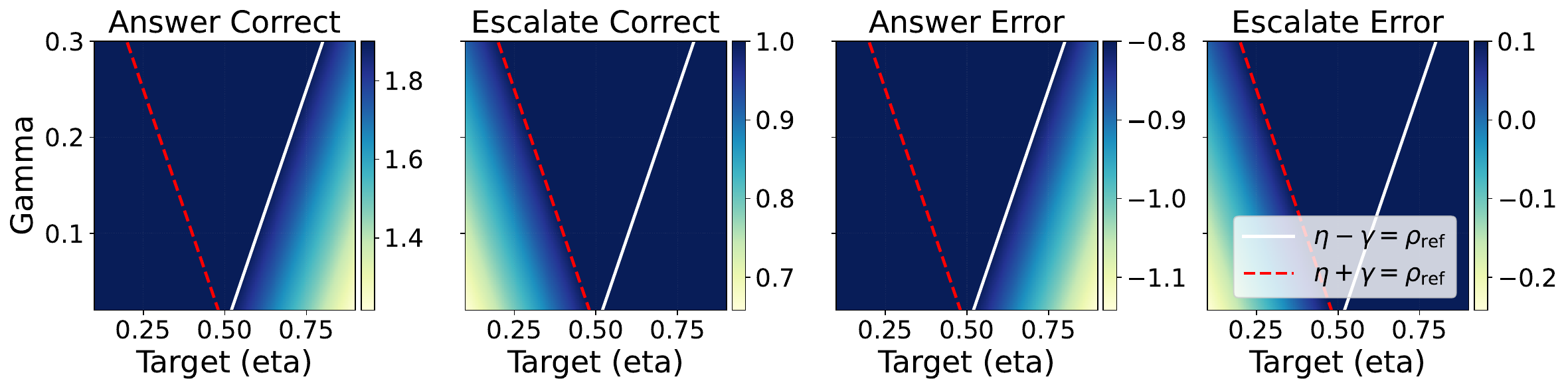}
  \caption{\textbf{Reward-surface visualization under target-band control.} With fixed $\rho_{\text{ref}}{=}0.5$ and format score $=0.1$, the four panels expand the piecewise target-band rule in Fig.~\ref{fig:tbgrpo_framework} into the \textit{Answer/Escalate} $\times$ \textit{Correct/Error} branches. The boundary lines indicate where answer-path or escalate-path penalties become active. This visualization serves as a sensitivity analysis for $(\eta,\gamma)$, while the main paper focuses Fig.~\ref{fig:tbgrpo_training_dynamics} on empirical training behavior.}
  \label{fig:supp_tbgrpo_target_gamma_effect}
\end{figure*}

Figure~\ref{fig:supp_tbgrpo_target_gamma_effect} shows how the controllable operating point changes with the target escalation ratio $\eta$ and tolerance band $\gamma$. Increasing $\eta$ shifts the preferred operating band toward higher escalation frequency, while increasing $\gamma$ widens the neutral region where neither branch receives a ratio penalty. This confirms that $(\eta,\gamma)$ gives TB-GRPO an explicit control interface over routing frequency, complementing the piecewise schematic in Fig.~\ref{fig:tbgrpo_framework}.

\subsection{Subtask-Level Escalation and Accuracy Analysis}

\begin{figure*}[t]
  \centering
  \includegraphics[width=\linewidth,height=0.84\textheight,keepaspectratio]{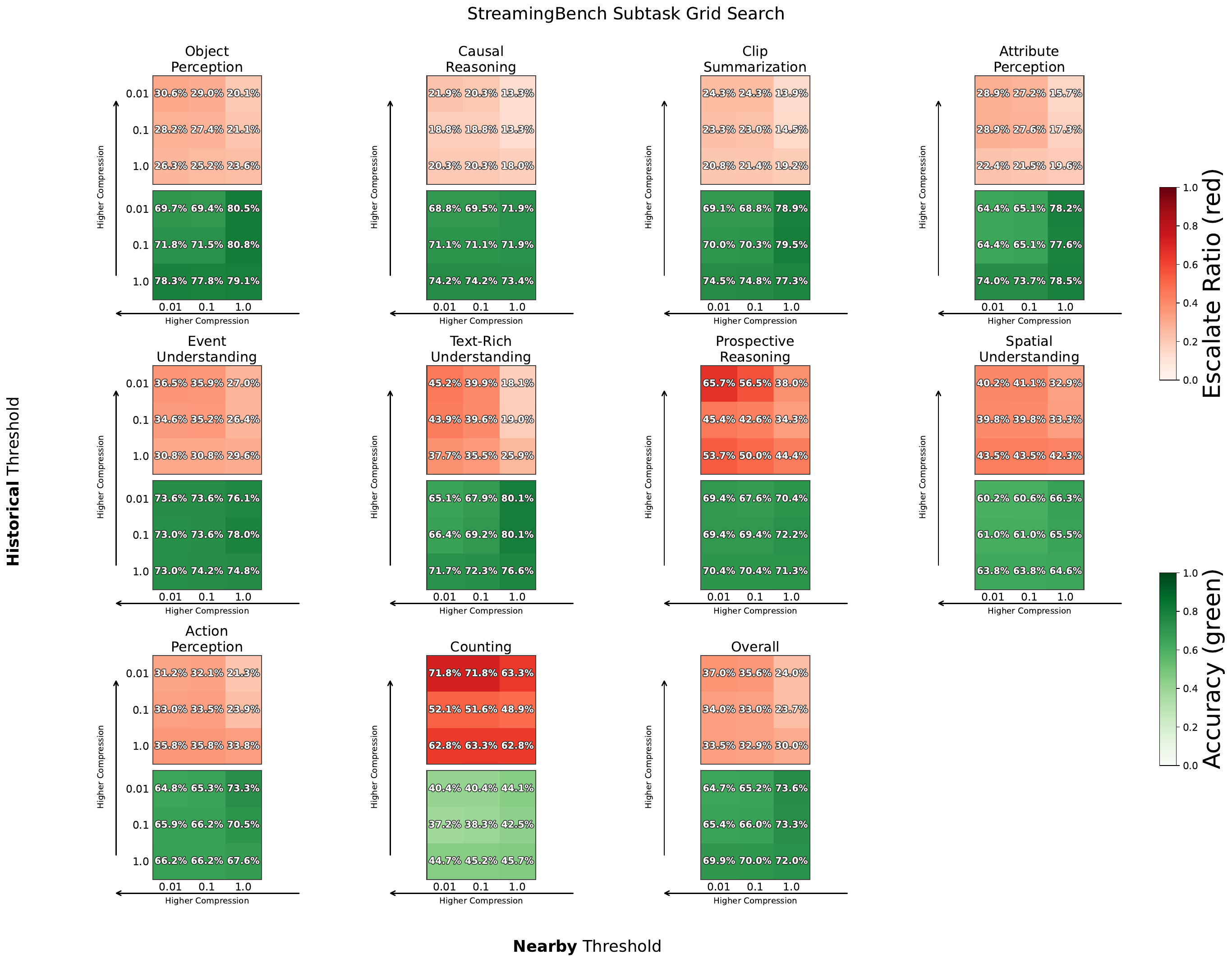}
  \caption{\textbf{StreamingBench subtask-level analysis of accuracy and escalation ratio.} Each panel displays the escalation ratio (top, red) and accuracy (bottom, green) under varying Historical and Nearby memory compression thresholds. Preserving recent evidence (Nearby=1.0) simultaneously boosts accuracy and naturally suppresses the need for slow-model escalation across most perception-oriented tasks (e.g., Object Perception). Conversely, cognitively demanding tasks like Prospective Reasoning and Counting consistently maintain higher escalation rates regardless of memory state, demonstrating the dynamic, task-driven nature of our \textit{Adaptive Thinking} routing policy.}
  \label{fig:supp_streamingbench_subtasks}
\end{figure*}

Figure~\ref{fig:supp_streamingbench_subtasks} provides a granular view of how our dynamic routing policy (Reason) behaves across different subtasks under varying memory compression states (Remember). The heatmaps show both the escalation ratio (red, top) and accuracy (green, bottom) for a comprehensive efficiency-performance analysis.

\subsubsection{Memory--Routing Synergy}
\label{app:memory route}
The clearest global pattern emerges in the ``Overall'' panel. Shifting from an overly aggressive compression state (Historical=0.01, Nearby=0.01) to our optimal age-aware policy (Historical=0.01, Nearby=1.0) significantly improves accuracy from 64.72\% to 73.64\%. Crucially, this accuracy boost is accompanied by a substantial decrease in the escalation ratio, dropping from 37.04\% to 24.04\%. This demonstrates a profound synergy between our modules: when the Remember module properly preserves nearby visual evidence, the fast model becomes highly capable of generating direct answers. Consequently, the optimal operating region is not achieved by blindly querying the slow model more often, but by supplying the fast path with high-fidelity recent context, thereby naturally suppressing unnecessary computational overhead.

\subsubsection{Task-Driven Routing Behaviors}
\label{app:task_route}
The subtask panels further reveal that our routing policy accurately adapts to intrinsic task difficulty. For perception-oriented tasks such as Object Perception, Attribute Perception, and Clip Summarization, restoring the Nearby Threshold to 1.0 makes the tasks both highly accurate and highly escalation-efficient. For example, Object Perception reaches 80.76\% accuracy while triggering the slow model only 20.05\% of the time. The fast model confidently handles these routine queries.

By contrast, cognitively demanding tasks like Counting and Prospective Reasoning inherently require deeper logic. As shown in their respective panels, even under optimal memory retention, Counting still necessitates a high escalation rate (ranging from 48.94\% to 71.81\%), and Prospective Reasoning maintains a baseline escalation rate above 65\%. This divergence perfectly validates the design of the Adaptive Thinking module via TB-GRPO: it dynamically allocates heavy reasoning budgets to anticipatory and complex reasoning tasks, while efficiently resolving perception-heavy queries on the fast path.

\section{Full Benchmark Results}
\label{sec:supp_full_results}

This section provides the complete subtask-level results for OVO-Bench and StreamingBench, expanding the category-level summaries in the main paper.

\subsection{Detailed OVO-Bench Results}

Table~\ref{tab:supp_main_ovobench_detail} expands the main OVO-Bench comparison in Table~\ref{tab:main_ovobench} by reporting every subcategory score. The main paper keeps only category-level averages to make the primary comparison easier to scan.

\begin{table*}[t]
\centering
\caption{\textbf{Detailed OVO-Bench comparison.} This table reports the full subcategory breakdown corresponding to Table~\ref{tab:main_ovobench}. For our entries, ``$96\%\downarrow$'' denotes the visual-token reduction ratio introduced by \textbf{Remember} (Active Forgetting) before downstream reasoning. Bold indicates the best result within the Streaming MLLMs block.}
\label{tab:supp_main_ovobench_detail}
\scriptsize
\setlength{\tabcolsep}{3pt}
\resizebox{\textwidth}{!}{
\begin{tabular}{l c c c c c c c c c c c c c c c c c}
\toprule
Model & \#Frames & \multicolumn{7}{c}{Real-Time Visual Perception} & \multicolumn{4}{c}{Backward Tracing} & \multicolumn{4}{c}{Forward Active Responding} & Overall Avg. \\
\cmidrule(lr){3-9}\cmidrule(lr){10-13}\cmidrule(lr){14-17}
& & OCR & ACR & ATR & STU & FPD & OJR & Avg. & EPM & ASI & HLD & Avg. & REC & SSR & CRR & Avg. & Overall Avg. \\
\midrule
\multicolumn{18}{l}{\textbf{Offline Video MLLMs}} \\
Qwen2-VL-72B \cite{qwen25vl2025} & 64 & 65.77 & 60.55 & 69.83 & 51.69 & 69.31 & 54.35 & 61.92 & 52.53 & 60.81 & 57.53 & 56.95 & 38.83 & 64.07 & 45 & 49.3 & 56.27 \\
Qwen3-VL-4B-Thinking \cite{qwen3vl2025} & 1fps & 72.48 & 55.05 & 71.55 & 46.07 & 74.26 & 58.15 & 62.93 & 50.17 & 66.89 & 37.63 & 51.56 & 45.20 & 69.36 & 61.67 & 58.74 & 57.74 \\
Qwen2.5-VL-7B \cite{qwen25vl2025} & 1fps & 76.51 & 61.47 & 69.83 & 53.37 & 68.32 & 58.70 & 64.70 & 51.85 & 62.84 & 19.35 & 44.68 & 34.63 & 44.09 & 46.67 & 41.80 & 50.39 \\
LLaVA-OneVision-7B \cite{llavaonevision2024} & 64 & 66.44 & 57.8 & 73.28 & 53.37 & 71.29 & 61.96 & 64.02 & 54.21 & 55.41 & 21.51 & 43.71 & 25.64 & 67.09 & 58.75 & 50.5 & 52.74 \\
Qwen2-VL-7B \cite{qwen25vl2025} & 64 & 60.4 & 50.46 & 56.03 & 47.19 & 66.34 & 55.43 & 55.98 & 47.81 & 35.48 & 56.08 & 46.46 & 31.66 & 65.82 & 48.75 & 48.74 & 50.39 \\
LongVU-7B \cite{longvu2024} & 1fps & 53.69 & 53.21 & 62.93 & 47.75 & 68.32 & 59.78 & 57.61 & 40.74 & 59.46 & 4.84 & 35.01 & 12.18 & 69.48 & 60.83 & 47.5 & 46.71 \\
\midrule
\multicolumn{18}{l}{\textbf{Streaming MLLMs}} \\
Flash-VStream-7B \cite{flashvstream2025} & 1fps & 24.16 & 29.36 & 28.45 & 33.71 & 25.74 & 28.8 & 28.37 & 39.06 & 37.16 & 5.91 & 27.38 & 8.02 & \textbf{67.25} & 60 & 45.09 & 33.61 \\
VideoLLM-online-8B \cite{videollmonline2024} & 2fps & 8.05 & 23.85 & 12.07 & 14.04 & 45.54 & 21.2 & 20.79 & 22.22 & 18.8 & 12.18 & 17.73 & - & - & - & - & - \\
Dispider-7B \cite{dispider2025} & 1fps & 57.72 & 49.54 & 62.07 & 44.94 & 61.39 & 51.63 & 54.55 & 48.48 & 55.41 & 4.3 & 36.06 & 18.05 & 37.36 & 48.75 & 34.72 & 41.78 \\
ET-Instruct-3B \cite{etinstruct2024} & 1fps & 65.1 & 35.78 & 56.9 & 35.39 & 24.75 & 60.87 & 46.47 & 41.81 & 35.14 & 8.6 & 28.52 & 20.06 & 52.31 & 67.5 & 46.62 & 40.54 \\
Streamo-3B \cite{streamo2025} & 1fps & 78.52 & 52.29 & 67.24 & 44.38 & 55.45 & \textbf{71.2} & 61.51 & 51.18 & 57.43 & 16.67 & 41.76 & 27.94 & 50.72 & 82.5 & 53.72 & 52.33 \\
Streamo-7B \cite{streamo2025} & 1fps & 79.19 & 57.8 & \textbf{75} & 49.44 & 64.36 & 70.11 & 65.98 & 54.55 & 52.03 & 31.72 & 46.1 & 29.96 & 51.03 & \textbf{83.33} & 54.77 & 55.61 \\
TimeChat-Online-7B (85\%$\downarrow$) \cite{timechatonline2025} & 1fps & 69.8 & 48.6 & 64.7 & 44.9 & 68.3 & 55.4 & 58.6 & 53.9 & 62.8 & 9.1 & 42 & 32.5 & 36.5 & 40 & 36.4 & 45.6 \\
StreamAgent-7B \cite{streamagent2025} & 1fps & 71.2 & 53.2 & 63.6 & 53.9 & 67.3 & 58.7 & 61.3 & 54.8 & 58.1 & 25.8 & 41.7 & 35.9 & 48.4 & 52 & 45.4 & 49.4 \\
FluxMem-7B \cite{fluxmem2026} & 1fps & - & - & - & - & - & - & - & - & - & - & - & - & - & - & - & 53.4 \\
R3-Streaming-3B|7B-Instruct (Ours, 96\%$\downarrow$) & 1fps & 83.22 & 58.72 & 63.79 & 55.06 & 66.34 & 63.59 & 65.12 & 57.91 & 64.19 & 11.83 & 44.64 & 34.24 & 66.01 & 57.08 & 52.44 & 54.07 \\
R3-Streaming-3B|4B-Thinking (Ours, 96\%$\downarrow$) & 1fps & 81.21 & 56.88 & 63.79 & 50.56 & 64.36 & 64.67 & 63.58 & \textbf{55.56} & \textbf{70.95} & 24.19 & 50.23 & 44.41 & 66.01 & 57.08 & \textbf{55.83} & 56.55 \\
R3-Streaming-7B|4B-Thinking (Ours, 96\%$\downarrow$) & 1fps & \textbf{87.92} & \textbf{70.64} & 73.28 & \textbf{59.55} & \textbf{69.32} & 70.65 & \textbf{71.89} & 51.18 & 65.54 & \textbf{37.1} & \textbf{51.27} & \textbf{45.13} & 57.09 & 49.58 & 50.60 & \textbf{57.92} \\
\bottomrule
\end{tabular}}
\end{table*}

\subsection{Detailed StreamingBench Results}

Table~\ref{tab:supp_main_streamingbench_detail} expands the main StreamingBench comparison in Table~\ref{tab:main_streamingbench} by reporting frame settings and all ten subtask scores. The main paper keeps only the official aggregate ``All'' score for compactness.

\begin{table*}[t]
\centering
\caption{\textbf{Detailed StreamingBench comparison.} This table reports the full subtask breakdown corresponding to Table~\ref{tab:main_streamingbench}. Bold indicates the best result within the Streaming MLLMs block, and ``All'' follows the official benchmark aggregation.}
\label{tab:supp_main_streamingbench_detail}
\scriptsize
\setlength{\tabcolsep}{3.2pt}
\resizebox{\textwidth}{!}{
\begin{tabular}{l c c c c c c c c c c c c c}
\toprule
Model & Params & Frames & OP & CR & CS & ATP & EU & TR & PR & SU & ACP & CT & All \\
\midrule
\multicolumn{14}{l}{\textbf{Human}} \\
Human & - & - & 89.47 & 92 & 93.6 & 91.47 & 95.65 & 92.52 & 88 & 88.75 & 89.74 & 91.3 & 91.46 \\
\midrule
\multicolumn{14}{l}{\textbf{Proprietary MLLMs}} \\
GPT-4o \cite{gpt4o2024} & - & 64 & 77.11 & 80.47 & 83.91 & 76.47 & 70.19 & 83.8 & 66.67 & 62.19 & 69.12 & 49.22 & 73.28 \\
Claude 3.5 Sonnet \cite{claude35sonnet2024} & - & 20 & 80.49 & 77.34 & 82.02 & 81.73 & 72.33 & 75.39 & 61.11 & 61.79 & 69.32 & 43.09 & 72.44 \\
\midrule
\multicolumn{14}{l}{\textbf{Offline Video MLLMs}} \\
LLaVA-OneVision \cite{llavaonevision2024} & 7B & 32 & 80.38 & 74.22 & 76.03 & 80.72 & 72.67 & 71.65 & 67.59 & 65.45 & 65.72 & 45.08 & 71.12 \\
Qwen2-VL \cite{qwen25vl2025} & 7B & 0.2-1 fps & 75.2 & 82.81 & 73.19 & 77.45 & 68.32 & 71.03 & 72.22 & 61.19 & 61.47 & 46.11 & 69.04 \\
Qwen3-VL-4B-Thinking \cite{qwen3vl2025} & 4B & 1 fps & 77.78 & 74.22 & 76.97 & 80.45 & 73.58 & 78.50 & 74.07 & 63.41 & 67.90 & 57.45 & 73.16 \\
Qwen2.5-VL-7B \cite{qwen25vl2025} & 7B & 1 fps & 80.22 & 79.69 & 79.18 & 81.41 & 72.96 & 77.26 & 76.85 & 62.20 & 65.06 & 53.19 & 73.28 \\
MiniCPM-V 2.6 \cite{minicpmv26_2024} & 8B & 32 & 71.93 & 71.09 & 77.92 & 75.82 & 64.6 & 65.73 & 70.37 & 56.1 & 62.32 & 53.37 & 67.44 \\
LLaVA-NeXT-Video \cite{llavavideo2024} & 32B & 64 & 78.2 & 70.31 & 73.82 & 76.8 & 63.35 & 69.78 & 57.41 & 56.1 & 64.31 & 38.86 & 66.96 \\
VILA-1.5 \cite{vila2023} & 8B & 14 & 53.68 & 49.22 & 70.98 & 56.86 & 53.42 & 53.89 & 54.63 & 48.78 & 50.14 & 17.62 & 52.32 \\
Video-CCAM \cite{videoccam2024} & 14B & 96 & 56.4 & 57.81 & 65.3 & 62.75 & 64.6 & 51.4 & 42.59 & 47.97 & 49.58 & 31.61 & 53.96 \\
Video-LLaMA2 \cite{videollama22024} & 7B & 32 & 55.86 & 55.47 & 57.41 & 58.17 & 52.8 & 43.61 & 39.81 & 42.68 & 45.61 & 35.23 & 49.52 \\
\midrule
\multicolumn{14}{l}{\textbf{Streaming MLLMs}} \\
Flash-VStream \cite{flashvstream2025} & 7B & - & 25.89 & 43.57 & 24.91 & 23.87 & 27.33 & 13.08 & 18.52 & 25.2 & 23.87 & 48.7 & 23.23 \\
VideoLLM-online \cite{videollmonline2024} & 8B & 2 fps & 39.07 & 40.06 & 34.49 & 31.05 & 45.96 & 32.4 & 31.48 & 34.16 & 42.49 & 27.89 & 35.99 \\
Dispider \cite{dispider2025} & 7B & 1 fps & 74.92 & 75.53 & 74.1 & 73.08 & 74.44 & 59.92 & 76.14 & 62.91 & 62.16 & 45.8 & 67.63 \\
TimeChat-Online-7B (83\%$\downarrow$) \cite{timechatonline2025} & 7B & 1 fps & 79.13 & \textbf{81.25} & 78.86 & 80.77 & 70.44 & 77.26 & 77.78 & 67.07 & 66.19 & 53.72 & 73.64 \\
StreamAgent \cite{streamagent2025} & 7B & 1 fps & 79.63 & 78.31 & 79.28 & 75.87 & 74.74 & 76.92 & 82.94 & 66.31 & \textbf{73.69} & \textbf{55.4} & 74.28 \\
R3-Streaming-3B|7B-Instruct (Ours, 95\%$\downarrow$) & 3B/7B & 1 fps & 81.03 & 75.78 & 79.18 & 77.88 & 77.36 & 78.82 & 74.07 & 64.23 & 71.31 & 48.40 & 73.84 \\
R3-Streaming-3B|4B-Thinking (Ours, 95\%$\downarrow$) & 3B/4B & 1 fps & 80.49 & 75.00 & 78.86 & 79.49 & 76.1 & 81.31 & 72.22 & 65.85 & 72.44 & 48.40 & 74.36 \\
R3-Streaming-7B|4B-Thinking (Ours, 95\%$\downarrow$) & 7B/4B & 1 fps & \textbf{82.38} & 77.34 & \textbf{84.23} & \textbf{82.37} & \textbf{80.5} & \textbf{82.87} & \textbf{84.26} & \textbf{69.92} & 69.03 & 43.62 & \textbf{76.36} \\
\bottomrule
\end{tabular}}
\end{table*}

\section{LLM Usage}
We acknowledge the use of a large language model (LLM) to assist in the preparation of this manuscript. The LLM's role was strictly limited to improving grammar and refining language. It did not contribute to any of the core research components, such as the initial ideas, experimental design, data analysis, or interpretation of the results.

\end{document}